\documentclass{article}

     \usepackage[preprint]{neurips_2020}

\usepackage{amsmath,amsfonts,bm}

\def\eqref#1{equation~\ref{#1}}

\def\1{\bm{1}}

\DeclareMathAlphabet{\mathsfit}{\encodingdefault}{\sfdefault}{m}{sl}
\SetMathAlphabet{\mathsfit}{bold}{\encodingdefault}{\sfdefault}{bx}{n}

\newcommand{\R}{\mathbb{R}}

\DeclareMathOperator*{\argmin}{arg\,min}

\usepackage[utf8]{inputenc} 
\usepackage[T1]{fontenc}    
\usepackage{adjustbox}
\usepackage{hyperref}       
\usepackage{url}            
\usepackage{booktabs}       
\usepackage{amsfonts}       
\usepackage{nicefrac}       
\usepackage{microtype}      
\usepackage{amsmath,amssymb,amsfonts}
\usepackage{algorithm}
\usepackage{listings}
\lstdefinestyle{ismll}{tabsize=4, basicstyle=\small\sffamily, columns=flexible,
 numbers=left, numbersep=5pt, numberstyle=\tiny, stepnumber=1, firstline=1, xleftmargin=8pt,
mathescape=True}
\usepackage{subfig}
\usepackage{color}
\usepackage{graphicx}
\usepackage{algorithmic}

\newcommand\commentout[1]{}
\title{Hyperparameter Optimization with Differentiable Metafeatures}

\author{%
  Hadi S.~Jomaa \\
  Department of Computer Science\\
  University of Hildesheim\\
  31141 Hildesheim, Germany \\
  \texttt{hsjomaa@ismll.de} \\
   \And
   Lars Schmidt-Thieme \\
  Department of Computer Science\\
  University of Hildesheim\\
  31141 Hildesheim, Germany \\
  \texttt{schmidt-thieme@ismll.de} \\
  \And
   Josif Grabocka \\
  Department of Computer Science\\
  University of Freiburg\\
  79110 Freiburg, Germany \\
  \texttt{grabocka@informatik.uni-freiburg.de} \\
}

\begin{document}

\maketitle

\begin{abstract}
Metafeatures, or dataset characteristics, have been shown to improve the performance of hyperparameter optimization (HPO). Conventionally, metafeatures are precomputed and used to measure the similarity between datasets, leading to a better initialization of HPO models. In this paper, we propose a cross dataset surrogate model called Differentiable Metafeature-based Surrogate (DMFBS), that predicts the hyperparameter response, i.e. validation loss, of a model trained on the dataset at hand. In contrast to existing models, DMFBS i) integrates a differentiable metafeature extractor and ii) is optimized using a novel multi-task loss, linking manifold regularization with a dataset similarity measure learned via an auxiliary dataset identification meta-task, effectively enforcing the response approximation for similar datasets to be similar. We compare DMFBS against several recent models for HPO on three large meta-datasets and show that it consistently outperforms all of them with an average 10\% improvement. Finally, we provide an extensive ablation study that examines the different components of our approach.
\end{abstract}
\section{Introduction}
\label{sec: intro}
Within the research community, the concentration of efforts towards solving the problem of hyperparameter optimization (HPO) has been mainly through sequential model-based optimization (SMBO). This process involves training a surrogate, typically a Gaussian process~(\cite{rasmussen2003gaussian}), to approximate the validation loss of a certain model trained with different hyperparameters, and suggesting the next hyperparameters via a policy, an acquisition function, that balances exploration and exploitation by leveraging the uncertainty in the posterior distribution~(\cite{jones1998efficient,wistuba2018scalable,snoek2012practical}). However, even when solutions are defined in conjunction with transfer learning techniques~(\cite{bardenet2013collaborative,wistuba2016two,feurer2015initializing}), the performance of SMBO solutions is heavily affected by the choice of the initial hyperparameters and can be improved by warm-start initialization~(\cite{bardenet2013collaborative,feurer2014using,feurer2015initializing,jomaa2019dataset2vec}).

In this paper, we present the problem of hyperparameter optimization as a meta-learning objective that exploits dataset information as part of the surrogate. Instead of treating HPO as a black-box function optimization problem, by operating blindly on the response of the hyperparameters alone, we treat it as a gray-box function~\cite{whitley2016gray} optimization problem, by capturing the relationship between the underlying dataset distribution and hyperparameters to better approximate the hyperparameter response function. 

We propose a novel formulation of the surrogate, which we call \textbf{M}eta\textbf{F}eature-\textbf{B}ased \textbf{S}urrogate (MFBS), that extends over the domain of metafeatures~(\cite{vanschoren2018meta}), i.e. dataset characteristics such as the number of instances, class probabilities, etc., Section~\ref{sec: mfbs}, and allows us to regress from the dataset metafeatures/hyperparameter pair directly onto the response. Driven by the assumption that similar datasets should have similar hyperparameter responses, we introduce a unique formulation of manifold regularization~(\cite{belkin2006manifold}) to penalize the difference between the \textit{approximated} hyperparameter response of datasets based on their metafeature similarity. 

As opposed to simply using precomputed metafeatures, we integrate a \textit{differentiable} (trainable) metafeature extractor into MFBS, resulting in a surrogate based on differentiable metafeatures, or DMFBS, and train the surrogate \textit{end-to-end}, constantly updating the metafeatures, Section~\ref{sec: dmfbs}.  DMFBS is initialized by meta-learning the initial parameters to approximate the response of a collection of datasets while explicitly learning to extract useful metafeatures.

We perform an extensive battery of experiments, Section~\ref{sec: experiments}, that highlight the transfer learning capacity of DMFBS by (1) outperforming the state-of-the-art HPO solutions for transfer learning and (2) performing an ablation study to analyze how the different components of the surrogate interact.

A summary of our contributions is:
\begin{enumerate}
	\item a formulation of DMFBS, a surrogate based on differentiable metafeatures that is meta-trained in an \textit{end-to-end} fashion to approximate the response of a collection of datasets and transferred to a target dataset by sequential fine-tuning;
	\item a novel multi-task optimization objective that links manifold regularization with a dataset similarity measure learned via an auxiliary dataset batch identification meta-task, effectively enforcing the response approximation for similar datasets to be similar;
	\item outperforming the state-of-the-art in HPO on a battery of experiments;
	\item three meta-datasets that, for the first time, include the associated datasets enabling novel solutions in HPO that involve metafeatures;	
	\item an extensive ablation study that highlights the importance of each component in our approach.
\end{enumerate}

\begin{figure}[h]
  \centering
  \includegraphics[width=0.5\columnwidth]{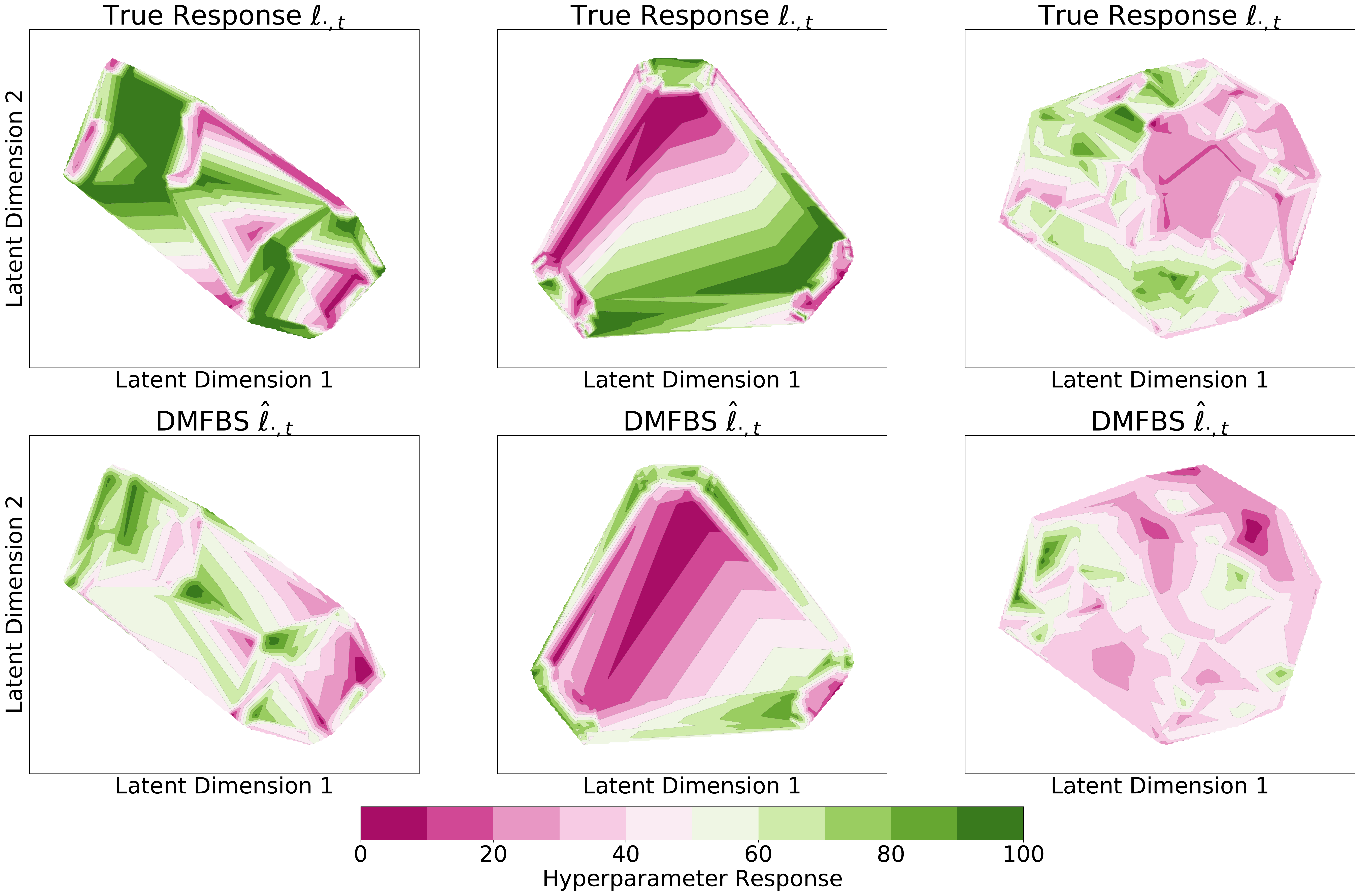}
  \caption{Actual response (top) and the approximated response (bottom) of hyperparameters for three datasets selected from the three meta-datasets. We reduce the dimensionality of each search space into a 2D representation via TSNE~(\cite{liu2016visualizing}). The optimal configurations have the least value.}
  \label{fig: zero-shot tsne}
\end{figure}

As a plausibility argument for the usefulness of DMFBS, we depict in Figure~\ref{fig: zero-shot tsne} the hyperparameter response surface for three datasets selected from the meta-test splits of the three meta-datasets, as well as the approximated hyperparameter response by DMFBS, after initializing its parameters via meta-learning, i.e. before any observations of hyperparameter responses for the target dataset at hand are made available.
\section{Related Work}
\label{related work}
A variety of methods have been proposed beyond the simple approaches~(\cite{bergstra2012random,brazdil2003ranking}), to leverage transfer learning for better HPO.

A common approach is warm-start initialization~(\cite{bardenet2013collaborative,feurer2014using,feurer2015initializing,jomaa2019dataset2vec}) of the surrogate based on dataset metafeature similarity~(\cite{rivolli2018towards}), which stems from the assumption that hyperparameters of similar datasets behave similarly to the response.
Transfer learning is also explored with the weighted combination of surrogates~(\cite{schilling2016scalable,wistuba2016two,feurer2018scalable}), or through learning a better initialization of the surrogate by training it jointly across different datasets~(\cite{perrone2018scalable,law2018hyperparameter,wistuba2021fewshot}).
Another prominent direction is learning a transferable acquisition function, which includes adjusting the score of the hyperparameters~(\cite{wistuba2018scalable}), or training a policy in a reinforcement learning setting (\cite{jomaa2019hyp,volpp2019meta}) to maximize a reward designed as a function of the regret.

An alternative direction revolves around the search space, such as applying pruning strategies~(\cite{wistuba2015hyperparameter}), or restricting it to the region where optimal hyperparameters are known to lie~(\cite{perrone2019learning}). Zero-shot HPO is also formulated as a stand-alone optimization problem, with different optimization objectives~(\cite{wistuba2015learning,wistuba2015sequential,winkelmolen2020practical}). However, these approaches cannot adapt to a target dataset and are outperformed by SMBO solutions with a very small number of observations.

In contrast to the literature, we formulate HPO as a gray-box function optimization problem, by designing a surrogate based on differentiable metafeatures. We rely on explicitly learned metafeatures to measure the similarity between datasets and penalize the difference between approximations of similar datasets via manifold regularization. This allows us to delineate from the complexity paired with Bayesian uncertainty and engineering similarity measures.

 \section{Hyperparameter Optimization}
\label{sec: problem formulation}
Let $\mathcal{D}$ be the space of all datasets and $\Lambda\subseteq\R^L$ the hyperparameter search space of a model under investigation, with $L$ as the number of hyperparameters. For example, one possible hyperparameter search space for a feedforward neural network might be $\Lambda:=\mathbb{N}\times\mathbb{N}\times\R^+_0$, where $L=3$ and the hyperparameters are the number of hidden layers, the number of neurons, and the learning rate, respectively. 

We denote by $\ell:\Lambda\times\cal{D}\rightarrow\R$
the function that yields for each dataset $D\in\cal{D}$ and hyperparameter $\lambda\in\Lambda$ the validation loss $\ell(\lambda,D)$ of the model trained on the training partition of the dataset $D$ for the given hyperparameter $\lambda$, and known as the \textit{hyperparameter response function}. We define $\hat{\ell}:\Lambda\times\mathcal{D}\rightarrow\R$ as the \textit{surrogate} for the true response function~(\cite{springenberg2016bayesian,law2018hyperparameter}), with $\theta$ as its parameters, that approximates the true hyperparameter response function.

The objective of hyperparameter optimization (HPO) is then to find the optimal hyperparameter $\lambda^* := \argmin_{\lambda\in\Lambda}\ell(\lambda,D)$ given a fixed budget $B$ of trials and $\mathcal{E}:= \left((D_1,\lambda_1,\ell_1),\dots,(D_N,\lambda_N,\ell_{N})\right)$ as a meta-dataset of primary datasets, hyperparameters and their responses from an unknown
  distribution $\rho_\mathcal{D}$ of datasets and an unknown distribution of hyperparameters, $\rho_\Lambda$\footnote{The triples represent the meta-dataset in a \textit{denormalized} way, instead of as grouped by the primary dataset. Thus, both, primary datasets and hyperparameters in general will occur in several triples.}.

Zero-shot HPO refers to the process of providing upfront the allowed number of hyperparameters to try without access to any hyperparameter response for the target dataset. 

Sequential HPO models, on the other hand, \textit{require} the availability of the hyperparameter response of previously evaluated hyperparameters on a target dataset.
Given an acquisition model, also known as an acquisition function~(\cite{movckus1975bayesian}), as $\hat{a}:\mathcal{D}\times\left(\Lambda\times\R\right)^*\rightarrow\Lambda$
, $\hat{a}$ is optimized to achieve the \textit{minimum} expected hyperparameter response, 
\begin{equation}
\argmin_{\hat{a}}\mathbb{E}_{D \sim \rho_\mathcal{D}} \ell\left(\left\{\lambda_1,\dots,\lambda_B\right\},D\right)
\end{equation}
where \begin{equation}\ell\left(\{\lambda_1,\dots,\lambda_B\},D\right):=\min_{b\in\{1,\dots,B\}}\ell\left(\lambda_b,D\right)
\end{equation} 
denotes the minimum (best) hyperparameter response of a set of tried hyperparameters. For a given dataset $D$ and a desired number $B$ of hyperparameters, the hyperparameters $\lambda$ are selected iteratively as,
\begin{equation}
\lambda_{b+1} := \hat{a}\left(D,((\lambda_1,\ell(\lambda_1,D)),\dots,(\lambda_b,\ell(\lambda_b,D)))\right)
\end{equation}
\commentout 
{
Given $\hat{\ell}$ as the surrogate, a separate model 
}

For single task sequential HPO, $\hat{a}$ depends only on the observations of losses of the target dataset, and contains explicitly a zero-shot model to yield the first $b$ hyperparameters, e.g. for $b=1$, $\lambda_1:=\hat{a}\left(D,\emptyset\right)$.  
\section{Greedy HPO}

Arguably the simplest HPO solution is an acquisition model $\hat{a}$ that \textit{greedily} selects hyperparameters based on the surrogate $\hat{\ell}$ from a set of $C$ hyperparameter candidates, $\Lambda_C\subseteq\Lambda$,
\begin{equation}
\label{eq: greedy}
\hat{a}_\text{greedy}\left(D,\Lambda_C,B,\hat{\ell}\right):=\argmin^B_{\lambda\in\Lambda_C}\hat{\ell}\left(\lambda,D\;; \theta\right)
\end{equation}
where $B\leq C$. Without a properly initialized surrogate however, $\hat{a}_\text{greedy}$ acts like random initialization.

The greedy HPO model is easy to lift to the sequential HPO problem in a generic and principled manner by i) iteratively updating $\hat{\ell}$ with the observed
hyperparameter responses of the target dataset $D$, and ii) greedily selecting new hyperparameters via Equation~\ref{eq: greedy}. 

\begin{algorithm}\caption{greedy-hpo}
\label{alg:greedy zero}
\begin{algorithmic}[1]
\STATE{\algorithmicrequire\;parameters $\theta$; hyperparameter candidates $\Lambda$; meta-dataset $\mathcal{E}$; learning rate $\eta$; target dataset $D$}
\STATE{\textbf{while} not converged \textbf{do}}
\STATE{~~~~~$\theta\leftarrow\textbf{update-model}\left(\mathcal{E},\theta,\eta,D\right)$}
\STATE{$\lambda \leftarrow \hat{a}_\text{greedy}\left(D,\Lambda,1,\hat{\ell}\left(\cdot;\theta\right)\right)$}
\STATE{\textbf{return} $\lambda$}
\end{algorithmic}
\end{algorithm}

\section{MFBS: Metafeature-based Surrogates}
\label{sec: mfbs}
One of the open challenges in HPO is how to regress the hyperparameter response on a dataset, s.t. $\hat{\ell}:\Lambda\times\mathcal{D}\rightarrow\R$, to capture the relationship between the dataset itself and the response. The issue resides in the fact that datasets are sets of instances with a different number of features and/or classes which makes them difficult to represent by a fixed-size vector, as opposed to hyperparameters.
We propose to represent datasets by suitable metafeatures considering that the use of engineered metafeatures has had a significant impact on HPO, particularly for initialization~(\cite{feurer2015initializing,schilling2016scalable}).

\subsection{Metafeature-based Regression}
Generally, we can understand a specific metafeature extractor (MFE) as a function $\phi: {\cal D}\rightarrow\R^K$ that describes any dataset $D$ by a concise vector of $K$ such metafeatures. 
The dataset-aware surrogate model
\begin{equation}
\hat\ell(\lambda,D):= \hat\ell^{\text{(MF)}}(\lambda,\phi(D))
\end{equation}
 then boils down to a regression model defined as $\hat{\ell}^\text{(MF)}:\Lambda\times\R^K\rightarrow\R$ that can be trained across datasets using a simple quadratic loss, 
\begin{equation}
\label{eq:loss-sur}
f^{\text{SUR}}(\theta^{\text{SUR}};{\cal E}) := \sum_{n=1}^N\left(\ell_{n} - \hat{\ell}_{n}\right)^2
\end{equation}

given $\hat{\ell}_n := \hat\ell^{\text{(MF)}}(\lambda_n,\phi_n; \theta^{\text{SUR}})$ and $\theta^{\text{SUR}}$ as the parameters of $\hat{\ell}^\text{(MF)}$. We denote  $\phi_n:=\phi(D_n)$ for simplicity of notation.
\subsection{Metafeature-based Manifold Regularization}
To stabilize models, especially when trained in low data regimes (semi-supervised setting), often manifold regularization~(\cite{belkin2006manifold}) is used, that explicitly enforces approximations for similar instances to be similar. Based on the assumption that hyperparameters of similar datasets behave similarly with respect to the response, we propose a regularization term that enforces this assumption as:
\begin{equation}
\label{eq:loss-mr}
f^{\text{MR}}(\theta^{\text{SUR}};{\cal E}) := \sum_{n=1}^{N-1}\sum_{m=n+1}^N \hat{s}_{n,m} \left(\hat{\ell}_{n} - \hat{\ell}_{m}\right)^2
\end{equation}

We denote by $\hat{s}_{n,m}$ a similarity measure between instances $\left(D_n,\lambda_n\right)$ and $\left(D_m,\lambda_m\right)$. We can use the dataset metafeatures to define a simple similarity measure between two datasets (\cite{jomaa2019dataset2vec}),
\begin{align}\label{eq:sim}
  \hat{s}(D_n,D_m) :=  e^{-\lVert\phi_n-\phi_m\rVert}
\end{align}
and exploit the fact that the hyperparameters $\lambda$ are discrete, e.g., on a grid, by defining a hyperparameter configuration to be similar only to itself, yielding the similarity measure  
for dataset / hyperparameter pairs as
\begin{align}
\hat{s}_{n,m} := & \hat{s}((D_n,\lambda_n),(D_m,\lambda_m))
  \\  := & \hat{s}(D_n, D_m)  \, \mathbb{I}(\lambda_n=\lambda_m)\notag
\end{align}

As a result, the smaller the Euclidean distance between the metafeatures of two datasets, \textit{for the same hyperparameter}, the higher the similarity measure $\hat{s}$. Thus the difference between the approximated response of similar datasets is emphasized, while that of dissimilar datasets is ignored.

\commentout{
\subsection{Training Objective}

The metafeature-based surrogate, MFBS, denoted by $\hat{\ell}^\text{(MF)}$ is trained by optimizing the objective:
\begin{eqnarray}
\label{eq:mf-based surrogate}
\theta^{*} := \argmin_{\theta} \;\; \mathcal{O}\left(\theta\right) + \alpha_\text{MR} \; \mathcal{R}\left(\theta\right)
\end{eqnarray}

where $\alpha_\text{MR}\in\R^+$ is the coefficient of manifold regularization. }

\section{Surrogates based on Differentiable Metafeatures}
\label{sec: dmfbs}

The extent to which it will be possible to regress the hyperparameter response
on the dataset metafeatures, depends heavily on these metafeatures
being expressive and suited for the task. Existing dataset metafeatures
such as those used by \cite{feurer2015initializing} (henceforth called \textbf{MF1})
or \cite{wistuba2016two} (\textbf{MF2}) deliver a fixed size vector
of length $46$ and $22$, respectively, 
through applying an engineered, fixed function $\phi$ on the dataset
that computes the number of instances, the number of features, etc.,
but can neither adapt to the underlying distribution of datasets nor
to the task of hyperparameter response approximation.

Therefore, we propose to use a parametrized, differentiable, and thus learnable
metafeature extractor $\phi$ such as Dataset2Vec (\cite{jomaa2019dataset2vec}), instead.
This metafeature extractor is implemented itself as a neural network, thus it
can be integrated into the surrogate model and learned end-to-end to extract those dataset metafeatures that are helpful for learning surrogate models.

Dataset2Vec represents datasets by all their predictor/target pairs and
encodes them using a deep set architecture~(\cite{zaheer2017deep}).

\subsection{Dataset2Vec: The Metafeature Extractor}
Each supervised (tabular) dataset $D := \left(x,y\right)$ consists of instances $x \in \mathcal{X}:=\R^{I\times F}$ and classes $y \in \mathcal{Y}:=\R^{I\times C}$ such that $I$, $F$ and $C$ represent the number of instances, features and classes respectively. The dataset can be further represented as a set of smaller components, \textit{set of sets}, $D := \bigcup_{f=1}^F\bigcup_{c=1}^C\left\{\left(\bigcup_{i=1}^I\{\left(x_{i,f},y_{i.c}\right)\}\right)\right\}$. A tabular dataset composed of columns (features, classes) and rows (instances) is reduced to ~\textit{single} predictor-target pairs instead of instance-target pairs. Based on this representation, the Dataset2Vec model has the structure 

\begin{align}
\label{dataset2vec}
\phi(D\;;\;\theta^\text{MFE}) = e_3 \left(\frac{1}{FC}\sum_{f=1}^{F}\sum_{c=1}^{C}e_2\left(\frac{1}{I}\sum_{i=1}^{I}e_1(x_{i,f},y_{i,c})\right)\right)
\end{align}
with
  $e_1: \R^2\rightarrow\R^{K_1}$,
  $e_2: \R^{K_1}\rightarrow\R^{K_2}$ and
  $e_3: \R^{K_2}\rightarrow\R^{K}$
represented by feedforward neural networks with $K_1$, $K_1$, and $K$ output units, respectively. We denote $\theta^\text{MFE}$ the parameters of $\phi$.

This architecture has several advantages: i) it is a set-based formulation that captures the correlation between each variable (predictor) and its assigned target and ii) is permutation-invariant, i.e. the output is unaffected by the ordering of the pairs in the set. Other set-based functions such as (\cite{edwards2016towards,lee2019set}) can also be used for metafeature extraction, however, we focus on this deep-set formulation~(\cite{jomaa2019dataset2vec}) because it already has been shown to perform well for dataset metafeature extraction and implementation is readily available.

Integrating both models, the surrogate model $\hat\ell$ and the dataset metafeature
extractor $\phi$ makes  objective $f^{\text{SUR}}$ and regularization function $f^{\text{MR}}$
  in Equations~\ref{eq:loss-sur} and \ref{eq:loss-mr} functions of both model parameters,
   $f^{\text{SUR}}(\theta^{\text{SUR}}, \theta^{\text{MFE}})$ and
   $f^{\text{MR}}(\theta^{\text{SUR}}, \theta^{\text{MFE}})$,
similar to Equations~\ref{eq:loss-sur} and \ref{eq:loss-mr}, but with  
\begin{align}
  \hat\ell_n := \hat\ell(\lambda_n, \phi(D_n; \theta^{\text{MFE}}); \theta^{\text{SUR}})
\end{align}
The overall model parameters then are $\theta:=(\theta^{\text{MFE}}, \theta^{\text{SUR}})$.

\subsection{The Auxiliary Dataset Batch Identification Task}
\label{auxiliary dataset id task}

While in principle it is possible to learn both models end-to-end, practically such an approach likely will not lead to promising results, simply because
meta-datasets for HPO are usually limited in size. Dataset2Vec foresees, for exactly this situation, to connect another auxiliary meta-task
with almost unlimited data that helps to extract useful metafeatures: dataset batch identification (DBI).


Here, a batch is a multi-fidelity subset of the dataset, i.e. a joint row and column sample, and the auxiliary meta-task at hand is to identify if two batches \textit{originate} from the same dataset or different ones.

Let $\mathcal{E}^{\text{DBI}}:=(D'_n,D''_n,s_n)_{n=1:N'}$ be a sampled meta-dataset of pairs of batches $\left(D'_n,D''_n\right)$ of the primary datasets and denote $s_n\in\{0,1\}$ as the label which indicates if both batches \textit{originate} from the same primary dataset or from different ones. Using the dataset similarity, Equation~\ref{eq:sim} as probability for originating from the same class, the negative loglikelihood loss is just
\begin{align}
f^{\text{DBI}}(\theta^{\text{MFE}}; {\cal E}^{\text{DBI}}) :=
  \sum_{n=1}^{N'}  s_n \log(\hat{s}_{n',n''})    \label{eq:loss-bid}
       + (1-s_n) \log (1-\hat{s}_{n',n''}))   \notag
\end{align}

\subsection{Training Objective}
Overall we train our surrogate model with differentiable metafeatures (DMFBS), to estimate the hyperparameter response and explicitly capture the dataset similarity by optimizing the following objective, \textit{end-to-end},

\begin{align}
  f( \theta; {\cal E},& {\cal E}^{\text{DBI}}) :=
       f^{\text{SUR}}(\theta^{\text{SUR}}, \theta^{\text{MFE}}; {\cal E})
  + \alpha_{\text{MR}} f^{\text{MR}}(\theta^{\text{SUR}}, \theta^{\text{MFE}}; {\cal E})
  +  \alpha_{\text{DBI}} f^{\text{DBI}}(\theta^{\text{MFE}}; {\cal E}^{\text{DBI}})
\end{align}
where $\alpha_{\text{MR}},\alpha_{\text{DBI}}\in\R^+_0$ and represent the coefficients of the manifold regularization and the auxiliary batch identification meta-task, respectively.
\begin{algorithm}\caption{update-model \newline(using $f^{SUR}$,$f^{MR}$ and $f^{DBI}$ from Equations~\ref{eq:loss-sur},~\ref{eq:loss-mr} and \ref{eq:loss-bid})}\label{alg:learn-dmfbs}
\begin{algorithmic}[1]
\STATE{\algorithmicrequire\;meta-dataset $\mathcal{E}$;parameters $\theta=(\theta^\text{SUR},\theta^\text{MFE})$; learning rate $\eta$; target dataset $D$}
\STATE{$D_n,\lambda_n,\ell_n \sim\text{Unif}\left(\mathcal{E}\mid D_n= D\right)$}
\STATE{$D_m,\lambda_m,\ell_m \sim\text{Unif}\left(\mathcal{E}\mid D_m \neq D\right)$}
\STATE{$D'_{n},D''_{n} \leftarrow \textbf{batch}(D_n),\textbf{batch}(D_n)$}
\STATE{$D'_{m} \leftarrow \textbf{batch}(D_m)$}
\STATE{Compute the gradients
		\begin{align*}
			g^{\text{SUR}}\leftarrow \nabla_{\theta^{\text{SUR}}}\bigl( &f^{\text{SUR}}(.; \{(D_n,\lambda_n,\ell_n)\})\\+\alpha_{\text{MR}} &f^{\text{MR}}(.; \{(D_n,\lambda_n,\ell_n),(D_m,\lambda_m,\ell_m) \})
              \bigl)
        \end{align*}
		\begin{align*}
			g^{\text{MFE}}\leftarrow \nabla_{\theta^{\text{MFE}}}\bigl( &f^{\text{MFE}}(.; \{(D_n,\lambda_n,\ell_n)\})\\+\alpha_{\text{MR}} &f^{\text{MR}}(.; \{(D_n,\lambda_n,\ell_n),(D_m,\lambda_m,\ell_m) \})
			\\+ \alpha_{\text{DBI}} &f^{\text{DBI}}(.; \{(D_n',D_n'',1)\\&~~~~~~~~~~,(D_n',D_m',\mathbb{I}(n=m))\}
              \bigl)
        \end{align*}}        
\STATE{Update parameters 
\begin{align*}
    &\theta^{\text{SUR}}\leftarrow \theta^{\text{SUR}} - \eta g^{\text{SUR}}\\
    &\theta^{\text{MFE}}\leftarrow \theta^{\text{MFE}} - \eta g^{\text{MFE}}
\end{align*}}
\STATE{\textbf{return} $(\theta^\text{SUR},\theta^\text{MFE})$}
\end{algorithmic}
\end{algorithm}

Please note that the model is learned from two meta-datasets:
  i) the one for the observed hyperparameter responses
     $\mathcal{E}:= \left((D_1,\lambda_1,\ell_1),\dots,(D_N,\lambda_N,\ell_{N})\right)$ and
   ii) the one for the batch identification task
     $\mathcal{E}^{\text{DBI}}:=(D'_n,D''_n,s_n)_{n=1:N'}$.
As any other multi-task model on multiple datasets, it can be learnt from
their cross product
      ${\cal E}^{\text{SUR}} \times {\cal E}^{\text{DBI}}$,
i.e., from randomly combined pairs, with standard learning algorithms,
e.g., stochastic gradient descent for neural networks. The
training algorithm should sample from the cross product like a generator, and
not materialize it, Algorithm~\ref{alg:learn-dmfbs}.


\subsection{Meta-learning DMFBS Initialization}
In the lack of a quantifiable uncertainty measure, HPO solutions that rely on a deterministic surrogate cannot leverage any exploration strategies, such as those available for Bayesian optimization solutions~(\cite{movckus1975bayesian,garivier2011upper,hennig2012entropy}), and must rely on the greedy acquisition function, Equation~\ref{eq: greedy}. As a remedy, we propose to learn a surrogate initialization based on meta-learning, such that the surrogate is initialized for the target dataset with previous knowledge about where the optimal hyperparameters might lie. 

Meta-learning has been established as an important approach to learning model initialization for fast adaptation into new domains, such that the initial solution resides on a local minimum, and can consequently quickly adapt to a target task with very little information, e.g few-shot learning and transferable architectures~(\cite{finn2017model,zoph2018learning,hospedales2020meta}). In the context of sequential HPO, it is essential to find the optimal hyperparameters with a very small budget, which can be interpreted as a variant of few-shot learning~(\cite{wistuba2021fewshot}).

In Algorithm~\ref{alg:pseudocode} we present the pseudo-code for meta-learning the initial parameters of DMFBS via the first-order meta-learning optimization routine~(\cite{nichol2018first}). In a meta-learning setting, the collection of datasets is divided into three partitions, $\mathcal{D}:=\mathcal{D}^\text{train}\cup\mathcal{D}^\text{valid}\cup\mathcal{D}^\text{test}$, with non-overlapping subsets.

\begin{algorithm}\caption{meta-learn-dmfbs-initialization}\label{alg:pseudocode}
\begin{algorithmic}[1]
\STATE{\algorithmicrequire\; meta-dataset $\mathcal{E}$; parameters $\theta$; learning rate $\eta$; inner update steps $v$; meta-batch size $n$}
\STATE{\textbf{while} not converged \textbf{do}}
\STATE{~~~~~~~$t_1,\dots,t_n\sim\text{Unif}\left(\{1,\dots,N\}\right)$}
\STATE{~~~~~~~\textbf{for} $i=1\text{ to }n$ \textbf{do}}
\STATE{~~~~~~~~~~~~~~$\theta_i\leftarrow\theta$}
\STATE{~~~~~~~~~~~~~~\textbf{for} $j=1\text{ to }v$ \textbf{do}}
\STATE{~~~~~~~~~~~~~~~~~~~~~$\theta_i\leftarrow$\textbf{update-model}$\left(\mathcal{E},\theta_i,\eta,D_{t_i}\right)$}
\STATE{~~~~~~~Update $\theta \leftarrow\theta + \eta\frac{1}{n}\sum_{i=1}^{n}\left(\theta_i-\theta\right)$}
\STATE{\textbf{return} $\theta$}

\end{algorithmic}
\end{algorithm}

Meta-learning DMFBS initialization allows DMFBS to quickly adapt to the target datasets with very few observations of the hyperparameter response. 

In Algorithm~\ref{alg: run-dmfbs}, we present DMFBS as a greedy HPO solution that is initialized via meta-learning. Note that initially, the meta-dataset $\mathcal{E}$ does not contain any observations for the target dataset.

\begin{algorithm}\caption{run-dmfbs}
\label{alg: run-dmfbs}
\begin{algorithmic}[1]
\STATE{\algorithmicrequire\;target dataset $D$; surrogate $\hat{\ell}$; hyperparameter search space $\Lambda$; meta-dataset $\mathcal{E}$; learning rate $\eta$; initial budget $b$; total budget $B$; inner update steps $v$; meta-batch size $n$}
\STATE{Initialize $\theta$ randomly}
\STATE{$\theta\leftarrow$\textbf{meta-learn-dmfbs-initialization}$\left(\mathcal{E},\theta,\eta,v,n\right)$ }
\STATE{$\Lambda_b\leftarrow\hat{a}_\text{greedy}\left(D,\Lambda,b,\hat{\ell}\left(\cdot;\theta\right)\right)$\COMMENT{\small Note $\Lambda_b=\{\lambda_i\}_{i=1:b}$}}
\STATE{$\mathcal{E}\leftarrow\mathcal{E}\cup\{\left(D,\lambda_i,\ell(\lambda_i,D)\right)\}_{i=1:b}$}
\STATE{\textbf{for} $i = b \text{ to } B$ \textbf{do}}
\STATE{~~~~~$\lambda \leftarrow \textbf{greedy-hpo}\left(\theta,\Lambda\setminus\Lambda_i,\mathcal{E},\eta,D\right)$}
\STATE{~~~~~$\mathcal{E}\leftarrow\mathcal{E}\cup\{\left(D,\lambda,\ell(\lambda,D)\right)\}$}
\STATE{~~~~~$\Lambda_{i+1}\leftarrow\Lambda_i\cup\{\lambda\}$}
\STATE{$\lambda^*\leftarrow\argmin_{\lambda\in\Lambda_B}\ell\left(\lambda,D\right)$}
\STATE{\textbf{return} $\lambda^*$}
\end{algorithmic}
\end{algorithm}

\subsection*{Network Architecture}
\label{subsection: network architecture}

\begin{figure}[ht]
  \centering
  \includegraphics[width=0.5\columnwidth]{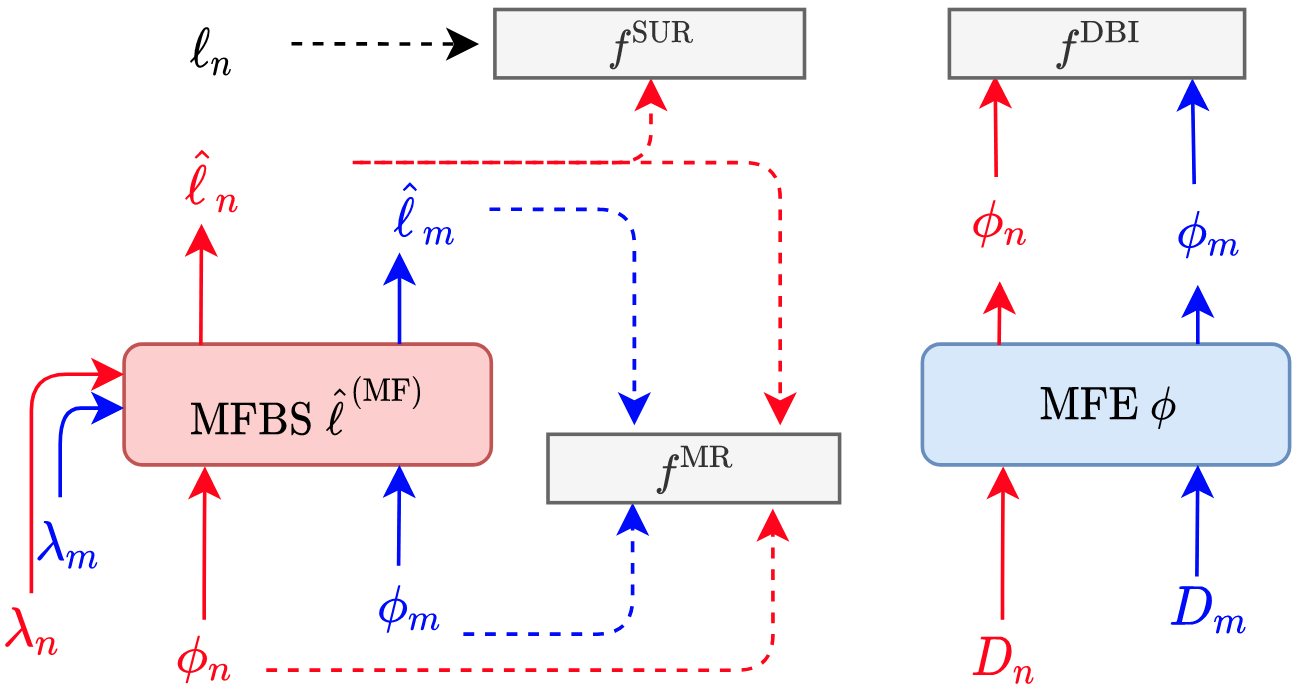}
  \caption{\textbf{DMFBS architecture} A pair of datasets, $(D_n,D_m)$, are processed by the same MFE $\phi$. Their respective metafeatures, $\phi_n$ and $\phi_m$, are paired with the hyperparameter $\lambda_n$ and $\lambda_m$, respectively, as the input to the surrogate $\hat{\ell}^\text{(MF)}$ to obtain $\hat{\ell}_{n}$ and $\hat{\ell}_{m}$. 
  }
  \label{fig: architectre}
\end{figure}

DMFBS is composed of two modules, $\hat{\ell}:= \phi\circ\hat{\ell}^\text{(MF)}$, namely the MFE $\phi$, and MFBS $\hat{\ell}^\text{(MF)}$, as depicted in Figure~\ref{fig: architectre}. The metafeature extractor $\phi:\R^2\rightarrow\R^{K}$ is composed of three functions, Equation~\ref{dataset2vec}, $\phi:e_1\circ e_2\circ e_3$.
MFBS is also composed of two functions, i.e. $\hat{\ell}^\text{(MF)}:\hat{\psi}_1\circ\hat{\psi}_2$. We define by $\hat{\psi}_1:\R^{K} \times \Lambda\rightarrow \R^{K_{\hat{\psi}_1}}$ as the function that takes as input the metafeature/hyperparameter pair, and by $\hat{\psi}_2:\R^{K_{\hat{\psi}_1}}\rightarrow\R$ the function that approximates the response.
Finally, let \textbf{Dense(n)} define one fully connected layer with $n$ neurons, and \textbf{ResidualBlock(n,m)} be $m\times$ Dense(n) with residual connections~(\cite{zagoruyko2016wide}). For all layers excluding the final one, we use a ReLU activation function. We present the details of the network architecture in Table~\ref{table: network architecture}.

\begin{table}[ht]
\caption{Network Architecture}
\centering
\scalebox{0.9}{
\begin{tabular}{ll}
\toprule
Functions &    Architecture\\
\midrule
$e_1$      &  Dense(32);$8\times$ResidualBlock(4,32);Dense(32)\\
$e_2$      &  $4\times$Dense(32)\\
$e_3$      &  Dense(32);$8\times$ResidualBlock(4,32);Dense(32)\\
$\hat{\psi}_1$ &  Dense(128);Dense(64);Dense(32);Dense(16)  \\
$\hat{\psi}_2$ & $4\times$Dense(16)\\
\bottomrule
\end{tabular}}
\label{table: network architecture}
\end{table}

\section{Experiments}
\label{sec: experiments}

Our experiments are designed to answer two research questions:
\begin{itemize}
  \item \textbf{Q1}:  Does meta-learning surrogates with end-to-end trainable metafeatures help generalize HPO on a new target dataset? 
  \item \textbf{Q2}: What is the impact of the manifold regularization and the dataset batch identification auxiliary tasks on the performance of transfer learning for HPO?  
\end{itemize}

\subsection{Meta-dataset}
A meta-dataset represents a collection of hyperparameters and their respective response, i.e. validation loss, for a certain model trained on several datasets, and is often created offline on a \textit{discretized} search space to expedite research in HPO. Existing meta-datasets~(\cite{feurer2015initializing,schilling2016scalable,perrone2019learning}) treat the original dataset as a nominal entity, and is therefore not provided to the community. However, the importance of the datasets is emphasized in this paper as they are necessary to generate metafeatures, and can be used as part of the surrogate domain. 

We present three meta-datasets by using 120 datasets chosen from the UCI repository~(\cite{asuncion2007uci}) and summarized in Table~\ref{table uci}. We then create the meta-instances by training a feedforward neural network and report the validation loss. Each dataset is provided with a predefined split 60\% train, 15\% validation, and 25\% test instances. We train each configuration for 50 epochs with a learning rate of $0.001$. The hyperparameter search space is described in Table~\ref{grid}.

The \textbf{layout} hyperparameter~(\cite{jomaa2019dataset2vec}) corresponds to the overall shape of the neural network and provides information regarding the number of neurons in each layer. For example, all the layers in the neural network with a $\square$ layout share the same number of \textbf{neurons}. We introduce an additional layout, $\bigtriangleup$, where the number of neurons in each layer is successively halved until it reaches the corresponding number of \textbf{neurons} in the central layer, then doubles successively.%

We also use dropout~(\cite{srivastava2014dropout}) and batch normalization~(\cite{ioffe2015batch}) as regularization strategies, and stochastic gradient descent (GD), ADAM~(\cite{DBLP:journals/corr/KingmaB14}) and RMSProp~(\cite{tieleman2012lecture}) as optimizers. SeLU~(\cite{klambauer2017self}) represents the self-normalizing activation unit.
The search space consists of all possible combinations of the hyperparameters. After removing redundant configurations, e.g. $\bigtriangleup$ layout with $1$ hidden layer and $16$ neurons is similar to $\square$ layout with with $1$ layer and $16$ neurons the resulting meta-datasets have $256$, $288$ and $324$ unique configurations respectively. 
\begin{table}[ht]
\caption{Hyperparameter search space for the meta-datasets. The name of each meta-dataset is inspired by the most prominent hyperparameter, highlighted in \textcolor{red}{red}.}
\centering
\centering
\scalebox{0.85}{
\begin{tabular}{llll}
\toprule
Hyperparameter &  Layout Md & Regularization Md & Optimization Md\\
\midrule
Activation   & ReLU, SeLU & ReLU, SeLU, LeakyReLU & ReLU, SeLU, LeakyReLU\\
Neurons   &  $4,8,16,32$ & $4,8,16,32$ & $4,8,16$ \\
Layers   & $1,3,5,7$ & $1,3,5,7$ & $3,5,7$ \\
Layout  &  \textcolor{red}{$\square$,$\lhd$,$\rhd$,$\diamond$,$\bigtriangleup$} & $\square$ & $\lhd$,$\rhd$,$\diamond$,$\bigtriangleup$ \\
Dropout&$0,0.5$ & \textcolor{red}{0, 0.2, 0.5}& $0$\\
Normalization& False & \textcolor{red}{False, True} & False\\
Optimizer& ADAM & ADAM & \textcolor{red}{ADAM, RMSProp, GD} \\
\bottomrule
\end{tabular}}
\label{grid}
\end{table}
\subsection*{Layout Hyperparameter}
Below are some examples of the number of neurons per layer for networks with different \textbf{layout} hyperparameters given 4 neurons and 5 layers:
\begin{itemize}
  \item Layout $\square$: [4,4,4,4,4]
  \item Layout $\lhd$: [4,8,16,32,64]
  \item Layout $\rhd$: [64,32,16,8,4]
  \item Layout $\diamond$: [4,8,16,8,4]
  \item Layout $\bigtriangleup$: [16,8,4,8,16]
\end{itemize}

The search space consists of all possible combinations of the hyperparameters. After removing redundant configurations, e.g. $\bigtriangleup$ layout with 1 layer is similar to a $\square$ layout with 1 layer, the resulting meta-datasets have $256$, $288$, and $324$ unique configurations respectively. 
\subsection*{Hyperparameter Encoding}
Below is the description of the encodings applied to our hyperparameters. We also like to note that the scalar values are normalized between $(0,1)$.

\begin{table}[ht]
\caption{Hyperparameter Encoding}
\centering
\centering
\scalebox{0.8}{
\begin{tabular}{ll}
\toprule
Hyperparameter & Encoding \\
\midrule
Activation   & One-hot encoding\\
Neurons   &  Scalar\\
Layers   & Scalar \\
Layout  &  One-hot encoding\\
Dropout& Scalar\\
Normalization& Scalar\\
Optimizer& One-hot encoding\\
\bottomrule
\end{tabular}}
\end{table}

\subsection{Training Protocol}
\label{sec: trainingprotocol}

In Algorithm~\ref{alg:pseudocode} we present the pseudo-code for meta-learning the initial parameters of DMFBS via the first-order meta-learning optimization routine~(\cite{nichol2018first}). We use the Adam~(\cite{DBLP:journals/corr/KingmaB14}) optimizer to optimize the inner model and stochastic gradient descent to optimize the parameters in the outer loop. We set the number of inner iterations to $v=5$, and use a learning rate of $\eta=0.01$ for both optimizers. For Adam, we set $\beta_1=0$ similar to the original meta-learning paper~(\cite{nichol2018first}). We use a batch size of $n=16$ tasks sampled uniformly at random. We also set the auxiliary loss coefficients $\alpha_\text{DBI}=0.1$ and $\alpha_\text{MR}=10$. We used the same hyperparameters and feedforward neural network architecture for all the variants. Finally, we perform $500$ outer iterations and stop when the \textit{greedy} performance for a budget $B=20$, Equation~\ref{eq: greedy}, on the datasets of the meta-valid set no longer improves. 

During fine-tuning the DMFBS to the new observations, i.e. after meta-learning, we use an Adam optimizer with the same learning rate, and the rest of the optimizer hyperparameters are set to their default values. We also use a batch size of $16$. In the case when the total number of observed hyperparameters is less than the batch size, we sample with repetition uniformly at random. The code is implemented in Tensorflow v2.4~(\cite{abadi2016tensorflow}).

\subsection{Evaluation Metrics}
The performance of black-box function optimizers is assessed by measuring the regret, representing the distance between an observed response and the optimal response on a response surface. For HPO, the meta-datasets are provided beforehand, consequently, the optimal response is known. We normalize the response surfaces between $(0,100)$, with $0$ being best, and thus observe the \textit{normalized} regret measured as $\min_{\lambda\in\Lambda_{B}}\ell\left(\lambda,D\right)$ with $\Lambda_{B}$ as the set of evaluated hyperparameters for target dataset $D$ after $B$ trials.

The reported results represent the average over 5-fold cross-validation split for each meta-dataset, with 80 meta-train, 16 meta-valid, and 24 meta-test datasets, and one unit of standard deviation.


\subsection{Results and Discussion}

\subsubsection*{\textbf{Q1:} Sequential HPO}
We compare DMFBS with the following HPO methods:
\begin{itemize}
  \item \textbf{Random} sampling~(\cite{bergstra2012random}) hyperparameters uniformly.
    \item \textbf{SMFO}~(\cite{wistuba2015sequential}) is a sequential model-free approach that provides an ordered sequence of hyperparameters by minimizing a ranking loss across all the tasks in the meta-train datasets.    
  \item \textbf{TST-R}~(\cite{wistuba2016two}) is a two-stage approach where the parameters of the target surrogate are adjusted via a kernel-weighted average based on the similarity of the hyperparameter response between the target dataset and the training datasets. We also evaluate the variant of this approach that relies on metafeatures, by replacing the engineered metafeatures with learned metafeatures, \textbf{TST-D2V}\footnote{Metafeatures found https://github.com/hadijomaa/dataset2vec}.
  \item \textbf{RGPE}~(\cite{feurer2018scalable}) is an ensemble model that estimates the target surrogate as a weighted combination of the training datasets' surrogates and the target surrogate itself. The weights are computed based on a ranking loss between the surrogates.
  \item \textbf{ABLR}~(\cite{perrone2018scalable}) is a multi-task ensemble of adaptive Bayesian linear regression models with all the datasets sharing a common feature extractor.
  \item \textbf{TAF-R}~(\cite{wistuba2018scalable}) is a transferable acquisition framework that selects the next hyperparameter based on a weighted combination of the expected improvement of the hyperparameters on the surrogate for target dataset, and the predicted improvement of the hyperparameter on the training datasets' surrogate.
  \item \textbf{MetaBO}~(\cite{volpp2019meta}) is a transferable acquisition function, optimized as a policy in a reinforcement learning framework. This approach, however, demands a pre-computed target surrogate as part of the state representation.
  \item \textbf{FSBO}~(\cite{wistuba2021fewshot}) presents HPO as a few-shot learning problem, where the a deep kernel network for a Gaussian process surrogate is
  trained to approximate the response of the training datasets, and used as an initialization strategy for the target dataset.
\end{itemize}
The results are summarized in Table~\ref{table: smbo}.

\begin{table}[h!]
\caption{Average normalized regret for HPO solutions for transfer learning. DMFBS consistently outperforms the baselines.  We report the best results in \textbf{bold} and \underline{underline} the second best. The results are reported every 33 trials.}
\centering
\begin{adjustbox}{angle=0,scale=0.76} 
\begin{tabular}{lccc}
\toprule
& \multicolumn{3}{c}{Layout Md} \\
Method &  @33 trials   &  @67 trials   &  @100 trials  \\
\toprule
Random &  5.221 $\pm$\footnotesize{  1.063} &  3.449 $\pm$\footnotesize{   0.690} &  2.225 $\pm$\footnotesize{   0.240}\\
SMFO      &  3.723 $\pm$\footnotesize{  0.957} &  2.400 $\pm$\footnotesize{  0.135} &  1.382 $\pm$\footnotesize{  0.576} \\
TST-R     &  3.582 $\pm$\footnotesize{  0.960} &  1.796 $\pm$\footnotesize{  0.767} &  1.140 $\pm$\footnotesize{  0.497} \\
TST-D2V   &  3.353 $\pm$\footnotesize{  0.424} &  1.725 $\pm$\footnotesize{  0.627} &  1.095 $\pm$\footnotesize{  0.573} \\
RGPE      &  \underline{2.637} $\pm$\footnotesize{  1.173} &  \underline{1.550} $\pm$\footnotesize{  1.001} &  1.132 $\pm$\footnotesize{  0.959} \\
ABLR      &  4.723 $\pm$\footnotesize{  0.801} &  2.536 $\pm$\footnotesize{  0.398} &  1.603 $\pm$\footnotesize{  0.690} \\
TAF-R     &  3.598 $\pm$\footnotesize{  0.835} &  2.725 $\pm$\footnotesize{  0.559} &  2.725 $\pm$\footnotesize{  0.559} \\
MetaBO    &  9.050 $\pm$\footnotesize{  2.235} &  8.629 $\pm$\footnotesize{  1.948} &  8.629 $\pm$\footnotesize{  1.948} \\
FSBO      &  4.159 $\pm$\footnotesize{  1.034} &  1.796 $\pm$\footnotesize{   0.740} &  \underline{0.974} $\pm$\footnotesize{  0.348} \\
\midrule
DMFBS-RI &  3.422 $\pm$\footnotesize{  0.938} &  1.969 $\pm$\footnotesize{  0.392} &  1.140 $\pm$\footnotesize{  0.303} \\
DMFBS &  \textbf{2.169} $\pm$\footnotesize{  0.369} &  \textbf{1.438} $\pm$\footnotesize{  0.437} &  \textbf{0.705} $\pm$\footnotesize{  0.232} \\
\toprule
\toprule
& \multicolumn{3}{c}{Regularization Md} \\
Method &  @33 trials   &  @67 trials   &  @100 trials  \\
\toprule
Random    &  6.044 $\pm$\footnotesize{  1.396} &  3.987 $\pm$\footnotesize{  1.002} &  3.014 $\pm$\footnotesize{  1.057} \\
SMFO      &  \underline{3.077} $\pm$\footnotesize{  0.688} &  1.995 $\pm$\footnotesize{  0.567} &  1.105 $\pm$\footnotesize{  0.595} \\
TST-R     &  3.738 $\pm$\footnotesize{  1.444} &  1.807 $\pm$\footnotesize{  0.741} &  1.033 $\pm$\footnotesize{  0.625} \\
TST-D2V   &  3.110 $\pm$\footnotesize{  1.304} &  1.581 $\pm$\footnotesize{  0.961} &  1.157 $\pm$\footnotesize{  0.832} \\
RGPE      &  \textbf{2.831} $\pm$\footnotesize{  1.265} &  \underline{1.397} $\pm$\footnotesize{  0.628} &  \underline{0.739} $\pm$\footnotesize{  0.307} \\
ABLR      &  4.771 $\pm$\footnotesize{  0.974} &  2.428 $\pm$\footnotesize{  0.837} &  1.618 $\pm$\footnotesize{  0.535} \\
TAF-R     &  3.416 $\pm$\footnotesize{  0.337} &  2.007 $\pm$\footnotesize{  1.183} &  2.007 $\pm$\footnotesize{  1.183} \\
MetaBO    &  8.865 $\pm$\footnotesize{  2.694} &  8.656 $\pm$\footnotesize{  2.500} &  8.093 $\pm$\footnotesize{  1.717} \\
FSBO      &  3.477 $\pm$\footnotesize{  0.950} &  2.436 $\pm$\footnotesize{  1.342} &  0.981 $\pm$\footnotesize{  0.732}  \\
\midrule
DMFBS-RI &  3.504 $\pm$\footnotesize{  1.791} &  2.464 $\pm$\footnotesize{  0.838} &  1.570 $\pm$\footnotesize{  0.715} \\
DMFBS &  3.079 $\pm$\footnotesize{  0.853} &  \textbf{1.153} $\pm$\footnotesize{  0.269} &  \textbf{0.669} $\pm$\footnotesize{  0.365} \\
\toprule
\toprule
& \multicolumn{3}{c}{Optimization Md} \\
Method &  @33 trials   &  @67 trials   &  @100 trials  \\
\toprule
Random    &  5.322 $\pm$\footnotesize{  1.074} &  4.081 $\pm$\footnotesize{  1.181} &  3.239 $\pm$\footnotesize{   0.880} \\
SMFO      &  3.875 $\pm$\footnotesize{  0.768} &  2.212 $\pm$\footnotesize{  0.460} &  1.591 $\pm$\footnotesize{  0.349} \\
TST-R     &  3.612 $\pm$\footnotesize{  0.950} &  2.132 $\pm$\footnotesize{  0.526} &  \underline{1.368} $\pm$\footnotesize{  0.481} \\
TST-D2V   &  3.939 $\pm$\footnotesize{  1.652} &  2.256 $\pm$\footnotesize{  0.545} &  1.455 $\pm$\footnotesize{  0.376} \\
RGPE      &  \underline{3.529} $\pm$\footnotesize{  1.127} &  \underline{1.682} $\pm$\footnotesize{  0.676} &  1.400 $\pm$\footnotesize{  0.556} \\
ABLR      &  7.255 $\pm$\footnotesize{  3.824} &  4.612 $\pm$\footnotesize{  2.349} &  2.246 $\pm$\footnotesize{  0.809} \\
TAF-R     &  4.608 $\pm$\footnotesize{  1.048} &  2.751 $\pm$\footnotesize{  0.821} &  2.747 $\pm$\footnotesize{  0.814} \\
MetaBO    &  7.638 $\pm$\footnotesize{  1.493} &  6.463 $\pm$\footnotesize{  2.411} &  6.141 $\pm$\footnotesize{  2.290} \\
FSBO      &  4.433 $\pm$\footnotesize{  1.130} &  2.159 $\pm$\footnotesize{  1.255} &  1.574 $\pm$\footnotesize{  0.433} \\
\midrule
DMFBS-RI &  4.076 $\pm$\footnotesize{  0.843} &  2.416 $\pm$\footnotesize{  0.821} &  1.375 $\pm$\footnotesize{  0.771} \\
DMFBS &  \textbf{3.198} $\pm$\footnotesize{  0.845} &  \textbf{1.614} $\pm$\footnotesize{  0.629} &  \textbf{1.323} $\pm$\footnotesize{  0.505} \\

\bottomrule
\end{tabular}
\end{adjustbox}
\label{table: smbo}
\end{table}
 DMFBS iteratively selects the hyperparameter with the highest score after being refit to the history of the observed hyperparameter responses on the target dataset. Contrary to the baselines that select hyperparameters through an acquisition function that capitalizes on the uncertainty of the posterior samples, DMFBS is a purely exploitative approach that demonstrates significant transfer learning capacity, consistently outperforming the state-of-the-art in HPO. Every sequential approach was initialized with one hyperparameter, followed by the sequential selection process.

The success of DMFBS in the lack of uncertainty estimates is heavily influenced by its initialization. DMFBS does not start the exploration from scratch, but it already has a prior-indication of where good local optima reside, considering that it had been meta-trained to approximate the response on a collection of datasets. That can be seen in Figure~\ref{fig: zero-shot tsne}. So exploration in the context of surrogates based on differentiable metafeatures is less critical than if HPO is treated entirely as a black-box problem because it already knows potential good local optima before fine-tuning. 

The effect of the initialization is evident when comparing with the randomly initialized DMFBS, referred to as DMFBS-RI. We notice that the performance significantly deteriorates. This shows the importance of proper initialization on the performance of DMFBS, which adapts quickly given very few observations of the response on the target dataset.

The advantage of DMFBS is further emphasized as we compare with TST-D2V, which uses learned metafeatures to measure dataset similarity as opposed to hyperparameter performance ranking features, as in TST-R and TAF-R. We notice once again that the use of learned metafeatures in TST-D2V outperforms on average the use of engineered metafeatures in both. 

We also notice MetaBO performs poorly since the reinforcement learning approach does not scale well with the increasing number of trials and hyperparameter search spaces. Finally, compared to another meta-learned initialization strategy, DMFBS outperforms FSBO, which requires a larger budget to overcome the rest of the baselines.

\subsubsection*{\textbf{Q2:} Ablation Study}
\label{section ablation}
We perform an ablation study to analyze the contribution of each component to the overall performance. While the main objective is to leverage sequential HPO, for the sake of analysis, we present the \textit{greedy} performance when no observations on the target dataset exist, i.e. after initialization based on meta-learning. 
\begin{table}[h]
\caption{Average normalized regret for MFBS and DMFBS optimized for different objectives. We report the best results in \textbf{bold} and \underline{underline} the second best.}
\centering
\scalebox{0.72}{
\begin{tabular}{lrrrrrr}
\toprule
& \multicolumn{3}{c}{Normalized Regret @5 trials} & \multicolumn{3}{c}{Normalized Regret @20 trials}\\
Method &    Layout Md &    Regularization Md &    Optimization Md &    Layout Md &    Regularization Md &    Optimization Md \\
\toprule
(Quadratic Loss)  \\
\midrule
MFBS(MF1)     &  13.146 $\pm$\footnotesize{  3.275 }&  11.418 $\pm$\footnotesize{  3.594 }&  11.586 $\pm$\footnotesize{ 3.413 }&  7.100 $\pm$\footnotesize{  2.219 }&  5.709 $\pm$\footnotesize{  1.964 }&  \underline{6.546} $\pm$\footnotesize{  1.817 }\\
MFBS(MF2)     &  11.755 $\pm$\footnotesize{  1.855 }&  13.175 $\pm$\footnotesize{  3.576 }&  12.997 $\pm$\footnotesize{ 3.887 }&  7.004 $\pm$\footnotesize{  1.461 }&  5.263 $\pm$\footnotesize{  0.828 }&  7.199 $\pm$\footnotesize{  2.649 }\\
MFBS(D2V)     &  11.441 $\pm$\footnotesize{  2.346 }&  11.673 $\pm$\footnotesize{  2.421 }&  12.930 $\pm$\footnotesize{ 4.121 }&  \underline{5.573} $\pm$\footnotesize{  1.986 }&  6.221 $\pm$\footnotesize{  2.140 }&  6.558 $\pm$\footnotesize{  1.703 }\\
DMFBS     &  \underline{10.952} $\pm$\footnotesize{  1.995} &  \textbf{10.077} $\pm$\footnotesize{  2.710}& 12.436 $\pm$\footnotesize{  3.543} & 5.952 $\pm$\footnotesize{  0.928} &  \underline{4.941} $\pm$\footnotesize{  1.196} &  6.786   $\pm$\footnotesize{  1.259}\\
\midrule
\multicolumn{3}{l}{(Quadratic Loss \& Manifold Regularization)}  \\
\midrule
MBFS(MF1)     &  13.238 $\pm$\footnotesize{  2.664 }&  12.061 $\pm$\footnotesize{  4.546 }&  11.769 $\pm$\footnotesize{ 2.926 }&  7.262 $\pm$\footnotesize{  1.813 }&  5.421 $\pm$\footnotesize{  1.264 }&  6.781 $\pm$\footnotesize{  1.586 }\\
MBFS(MF2)     &  12.499 $\pm$\footnotesize{  3.042 }&  11.293 $\pm$\footnotesize{  4.346 }&  12.660 $\pm$\footnotesize{ 4.554 }&  6.223 $\pm$\footnotesize{  1.219 }&  5.671 $\pm$\footnotesize{  1.364 }&  6.881 $\pm$\footnotesize{  2.494 }\\
MBFS(D2V)     &  12.367 $\pm$\footnotesize{  3.874 }&  12.044 $\pm$\footnotesize{  3.749 }&  \textbf{10.328} $\pm$\footnotesize{ 1.684 }&  6.596 $\pm$\footnotesize{  1.167 }&  6.660 $\pm$\footnotesize{  2.490 }&  6.621 $\pm$\footnotesize{  1.739 }\\
DMBFS     &  11.388 $\pm$\footnotesize{  2.710} &  \underline{10.536} $\pm$\footnotesize{  1.411}&  11.615 $\pm$\footnotesize{  1.629} &  5.662 $\pm$\footnotesize{  1.737} &  5.025 $\pm$\footnotesize{  0.948} &  6.952 $\pm$\footnotesize{  1.178}\\
\midrule
\multicolumn{5}{l}{(Quadratic Loss \& Manifold Regularization \& Dataset Batch Identification)} \\
\midrule
DMFBS      &  \textbf{10.605} $\pm$\footnotesize{  2.627 }&  10.998 $\pm$\footnotesize{  2.413 }&  \underline{10.451} $\pm$\footnotesize{ 1.956  }&  \textbf{5.263} $\pm$\footnotesize{  1.389 }&  \textbf{4.857} $\pm$\footnotesize{  1.406 }&  \textbf{5.548} $\pm$\footnotesize{  0.243 }\\
\bottomrule
\end{tabular}}
\label{table: ablation}
\end{table}
To further understand the source for the empirical gain of our method against the baselines, we compare DMFBS with MFBS trained with different metafeatures, namely: \textbf{MF1}~(\cite{feurer2015initializing}), \textbf{MF2}~(\cite{wistuba2016two}), and learned metafeatures \textbf{D2V}~(\cite{jomaa2019dataset2vec}). To highlight the contribution of manifold regularization and the auxiliary dataset batch identification meta-task, we juxtapose the results achieved when using different optimization objectives to meta-learn the initial parameters, Table~\ref{table: ablation}. We report the results for the first $20$ trials~(\cite{feurer2015initializing}) due to our interest in the early gains that can be achieved by each surrogate before fine-tuning.

As a summary of the ablation study, we conclude that:
\begin{enumerate}
  \item meta-learning the initial parameters using the quadratic loss as the sole objective works best with DMFBS, highlighting the importance of differentiable metafeatures;
  \item the effect of manifold regularization is dictated by the metafeatures. Coupling the quadratic loss with manifold regularization to meta-learn the initial parameters of DMFBS adversely affects the performance in the absence of DBI, the same applies when using fixed metafeatures. Nevertheless, DMFBS is better than the MFBS variants;
  \item introducing DBI improves the quality of the metafeatures, i.e. generates better representations of datasets, which consequently improves manifold regularization, leading to better initialization.
\end{enumerate}

\section{Conclusion}
In this paper, we formulate HPO as a gray-box function optimization problem that incorporates the dataset in the surrogate model. Specifically, we design a novel surrogate based on differentiable metafeatures, that is initialized through meta-learning and thus can adapt quickly to new target datasets with little observations. We propose and optimize a novel multi-task objective that links manifold regularization with a similarity measure based on the metafeatures that are in turn learned in end-to-end manner. As a result, we outperform the state-of-the-art in sequential HPO for transfer learning. 

\bibliography{2019d-jomaa}
\bibliographystyle{spbasic}

\appendix
\clearpage
\section{Additional Experiments}
\label{app: trainingprotocol}

\subsubsection*{\textbf{Q1:} Sequential HPO}

\begin{figure}[ht]
  \centering
  \includegraphics[width=0.9\columnwidth]{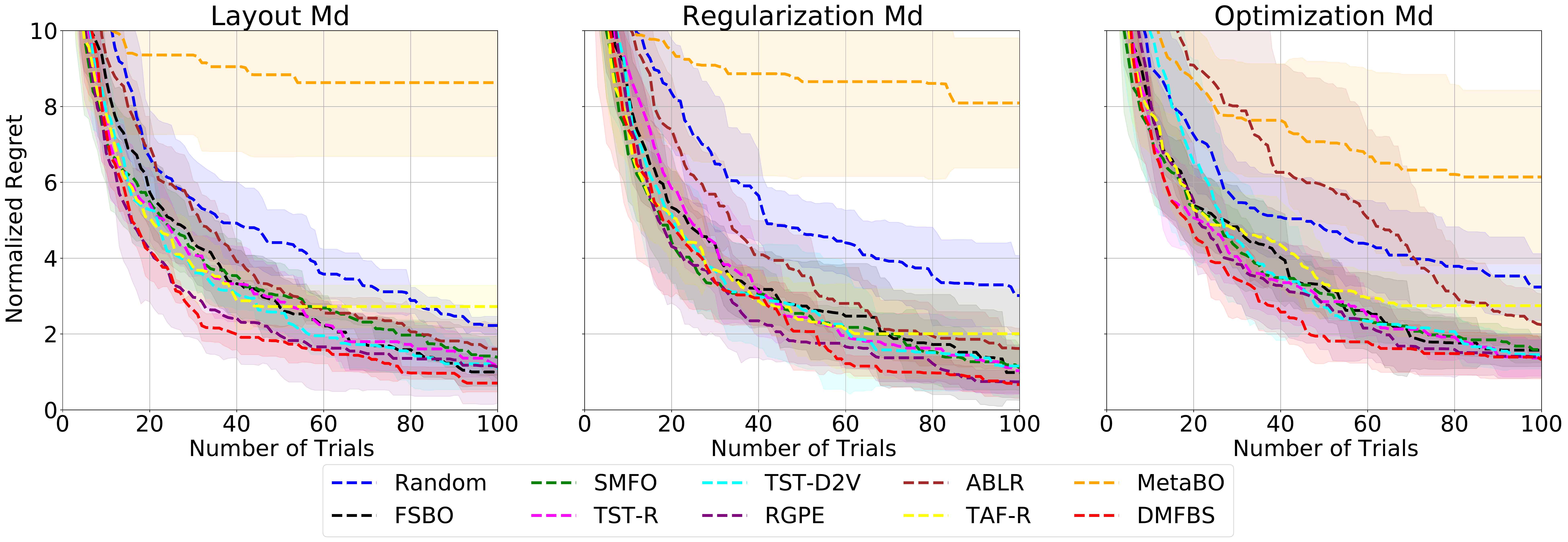}
  \caption{Average normalized regret for sequential HPO solutions for transfer learning. The shaded region represents one unit of standard deviation.} 
  \label{fig: smbo}
\end{figure}

We presented in Table~\ref{table: smbo} the aggregated performance every $33$ trials. In Figure ~\ref{fig: smbo} we show results for all $100$ trials.

\subsubsection*{\textbf{Q3:} DMFBS at Initialization for Zero-shot HPO}
\label{zero shot section}
Zero-shot HPO is the process of providing upfront a set of hyperparameters to try, without access to any observations of losses on the target dataset. Since we have shown that by meta-learning the parameters of DMFBS, we can quickly adapt to observations from the target datasets, we would like to examine the performance in case none exists. For comparison, we introduce the following baselines:

\begin{itemize}
    \item \textbf{Random} sampling~(\cite{bergstra2012random}).
    \item \textbf{Average Rank} presents the hyperparameters in an ordered sequence based on their average ranking across the meta-train datasets.
    \item \textbf{NN(METAFEATURE)}~(\cite{feurer2015initializing}) is the process of selecting the best hyperparameters of the nearest neighboring dataset based on the metafeature similarity. We use two sets of well-established engineered metafeatures, which we refer to as \textbf{MF1}~(\cite{feurer2015initializing}) and \textbf{MF2}~(\cite{wistuba2016two}), as well as learnt metafeatures\footnote{https://github.com/hadijomaa/dataset2vec}, \textbf{D2V}~(\cite{jomaa2019dataset2vec}). The similarity is measured by the Euclidean distance.
    \item \textbf{Ellipsoid}~(\cite{perrone2019learning}) is a random sampling approach, however, the hyperparameters are sampled from a hyper-ellipsoid search space that is restricted to encompass as many optimal hyperparameters from the training dataset as possible.  

\end{itemize}

In Table~\ref{table: zero-shot} we report the normalized regret achieved by the different approaches for the first $5$ and $20$ trials. 
As seen in~(\cite{jomaa2019dataset2vec}), learned metafeatures serve as a better representation of datasets compared to engineered metafeatures, leading to better performance, whereas sampling from the restricted hyper-ellipsoid outperforms the strategies with precomputed metafeatures.

Compared to the baselines, we notice that DMFBS at initialization outperforms sampling-based and metafeature-based strategies. As opposed to learning metafeatures in a meta-agnostic setting, the metafeature extractor is trained to capture the relationship between the dataset and the hyperparameter response directly. As a result, given an unseen dataset, DMFBS provides dataset-conditioned scores for the hyperparameters, which are in turn selected in the order of their best score. 

\begin{table}[t]
\caption{Average normalized regret for HPO initialization strategies compared with DMFBS at initialization. We report the best results in \textbf{bold} and \underline{underline} the second best.}
\centering
\scalebox{0.75}{
\begin{tabular}{lrrrrrr}
\toprule
& \multicolumn{3}{c}{Normalized Zero-shot Regret @5 Trials} & \multicolumn{3}{c}{Normalized Zero-shot Regret @20 Trials}\\
\midrule
Method &    Layout Md &    Regularization Md &    Optimization Md &    Layout Md &    Regularization Md &    Optimization Md \\
\toprule
Random     &  13.752 $\pm$\footnotesize{  2.496 }&  14.931 $\pm$\footnotesize{  2.303 }&  13.260 $\pm$\footnotesize{ 1.781  }&  6.664 $\pm$\footnotesize{  1.238 }&  8.351 $\pm$\footnotesize{  1.424 }&  7.250 $\pm$\footnotesize{  1.381 }\\
Average Rank&  13.103 $\pm$\footnotesize{  2.679 }&  \underline{10.999} $\pm$\footnotesize{  1.974 }&  12.298 $\pm$\footnotesize{ 2.683  }&  6.660 $\pm$\footnotesize{  1.947 }&  5.152 $\pm$\footnotesize{  1.059 }&  7.076 $\pm$\footnotesize{  2.255 }\\
NN(MF1)       &  14.611 $\pm$\footnotesize{  1.507 }&  14.098 $\pm$\footnotesize{  2.472 }&  12.770 $\pm$\footnotesize{ 1.110  }&  6.407 $\pm$\footnotesize{  1.068 }&  6.775 $\pm$\footnotesize{  1.846 }&  6.860 $\pm$\footnotesize{  1.044 }\\
NN(MF2)       &  12.981 $\pm$\footnotesize{  1.450 }&  13.600 $\pm$\footnotesize{  2.115 }&  12.740 $\pm$\footnotesize{ 1.112  }&  6.578 $\pm$\footnotesize{  0.719 }&  7.143 $\pm$\footnotesize{  0.964 }&  7.026 $\pm$\footnotesize{  1.210 }\\
NN(D2V)        &  11.733 $\pm$\footnotesize{  2.567 }&  11.959 $\pm$\footnotesize{  3.132 }&  \underline{11.173} $\pm$\footnotesize{ 2.200  }&  \underline{5.906} $\pm$\footnotesize{  0.865 }&  5.899 $\pm$\footnotesize{  1.759 }&  6.650 $\pm$\footnotesize{  1.357 }\\
Ellipsoid &  \underline{11.240} $\pm$\footnotesize{  2.314 }&  11.289 $\pm$\footnotesize{  3.161 }&  11.341 $\pm$\footnotesize{ 1.085  }&  6.070 $\pm$\footnotesize{  0.593 }&  \underline{4.930} $\pm$\footnotesize{  1.106 }&  \underline{6.606} $\pm$\footnotesize{  1.155 }\\

\midrule
DMFBS      &  \textbf{10.605} $\pm$\footnotesize{  2.627 }&  \textbf{10.998} $\pm$\footnotesize{  2.413 }&  \textbf{10.451} $\pm$\footnotesize{ 1.956  }&  \textbf{5.263} $\pm$\footnotesize{  1.389 }&  \textbf{4.857} $\pm$\footnotesize{  1.406 }&  \textbf{5.548} $\pm$\footnotesize{  0.243 }\\
\bottomrule
\end{tabular}}
\label{table: zero-shot}
\end{table}

\subsubsection*{\textbf{Q4:} DMFBS as an initialization strategy for single-task sequential HPO}
\begin{figure}[ht]
  \centering
  \includegraphics[width=0.9\columnwidth]{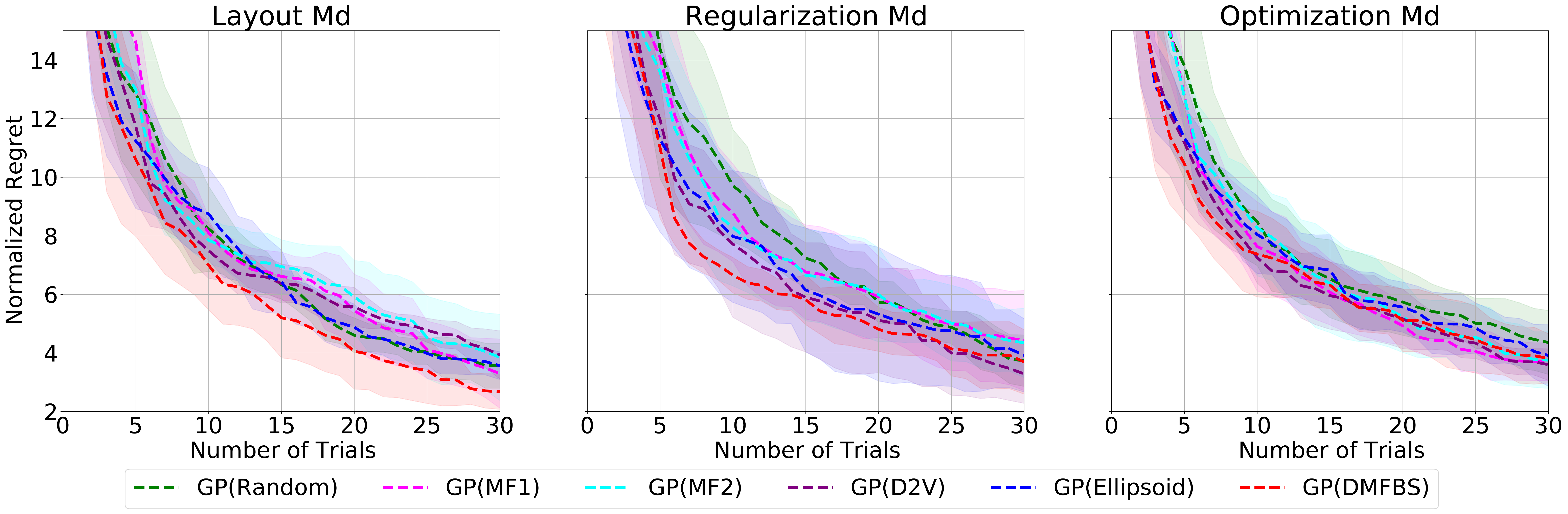}
  \caption{Average normalized regret for single-task sequential model-based optimization based on a Gaussian process surrogate with different initialization strategies, in parenthesis.} 
  \label{fig: gp warm}
\end{figure}

We use the aforementioned initialization strategies to warm-start single task GP~(\cite{rasmussen2003gaussian}) with an RBF kernel and automatic relevance determination as the surrogate, with $5$ hyperparameters. The quality of the hyperparameters selected based on their assigned score using DMFBS is reflected in the improved performance of the surrogate at the early stages compared to metafeature-based initialization and random sampling, Figure~\ref{fig: gp warm}.

\begin{table*}[ht]
\caption{Summary of the 120 UCI datasets used to generate the meta-datasets.}
\label{table uci}
\centering
\scalebox{0.8}{
\begin{adjustbox}{angle=0} 
\begin{tabular}{lrrrlrrr}
\toprule
                     UCI Dataset &  \# Instances &    \# Features &    \# Classes &                           UCI Dataset &  \# Instances &    \# Features &    \# Classes \\
\midrule
         molec-biol-splice &         2393 &   60 &    3 &                           adult &        32561 &   14 &    2 \\
                   twonorm &         5550 &   20 &    2 &                       annealing &          798 &   31 &    5 \\
               plant-texture &         1199 &   64 &  100 &             molec-biol-promoter &           80 &   57 &    2 \\
                    ringnorm &         5550 &   20 &    2 &                         contrac &         1105 &    9 &    3 \\
                       spect &           79 &   22 &    2 &                 statlog-landsat &         4435 &   36 &    6 \\
                   energy-y2 &          576 &    8 &    3 &    conn-bench-sonar-mines-rocks &          156 &   60 &    2 \\
                steel-plates &         1456 &   27 &    7 &                          musk-2 &         4949 &  166 &    2 \\
    vertebral-column-3clases &          233 &    6 &    3 &                        balloons &           12 &    4 &    2 \\
                  chess-krvk &        21042 &    6 &   18 &                         abalone &         3133 &    8 &    3 \\
             statlog-shuttle &        43500 &    9 &    7 &                 statlog-vehicle &          635 &   18 &    4 \\
          breast-cancer-wisc &          524 &    9 &    2 &                     page-blocks &         4105 &   10 &    5 \\
                     semeion &         1195 &  256 &   10 &                 heart-hungarian &          221 &   12 &    2 \\
                   connect-4 &        50668 &   42 &    2 &                      ionosphere &          263 &   33 &    2 \\
                     monks-3 &          122 &    6 &    2 &               synthetic-control &          450 &   60 &    6 \\
              wall-following &         4092 &   24 &    4 &                     plant-shape &         1200 &   64 &  100 \\
    vertebral-column-2clases &          233 &    6 &    2 &      pittsburg-bridges-MATERIAL &           80 &    7 &    3 \\
                    planning &          137 &   12 &    2 &         breast-cancer-wisc-diag &          427 &   30 &    2 \\
    cardiotocography-3clases &         1595 &   21 &    3 &                          spectf &           80 &   44 &    2 \\
                plant-margin &         1200 &   64 &  100 &                            bank &         3391 &   16 &    2 \\
                     nursery &         9720 &    8 &    5 &                       pendigits &         7494 &   16 &   10 \\
                     titanic &         1651 &    3 &    2 &                        teaching &          113 &    5 &    3 \\
                   energy-y1 &          576 &    8 &    3 &                        mushroom &         6093 &   21 &    2 \\
                     monks-1 &          124 &    6 &    2 &                         optical &         3823 &   62 &   10 \\
                  arrhythmia &          339 &  262 &   13 &                   primary-tumor &          248 &   17 &   15 \\
               breast-tissue &           80 &    9 &    6 &      conn-bench-vowel-deterding &          528 &   11 &   11 \\
   statlog-australian-credit &          518 &   14 &    2 &                         soybean &          307 &   35 &   18 \\
                 tic-tac-toe &          719 &    9 &    2 &    oocytes\_merluccius\_states\_2f &          767 &   25 &    3 \\
                lymphography &          111 &   18 &    4 &                     chess-krvkp &         2397 &   36 &    2 \\
                     monks-2 &          169 &    6 &    2 &                   audiology-std &          171 &   59 &   18 \\
                    waveform &         3750 &   21 &    3 &              image-segmentation &          210 &   18 &    7 \\
                   fertility &           75 &    9 &    2 &                     led-display &          750 &    7 &   10 \\
                      lenses &           18 &    4 &    3 &                        heart-va &          150 &   12 &    5 \\
            wine-quality-red &         1199 &   11 &    6 &          pittsburg-bridges-SPAN &           69 &    7 &    3 \\
                  parkinsons &          146 &   22 &    2 &  oocytes\_trisopterus\_nucleus\_2f &          684 &   25 &    2 \\
          wine-quality-white &         3674 &   11 &    7 &           statlog-german-credit &          750 &   24 &    2 \\
                        pima &          576 &    8 &    2 &              acute-inflammation &           90 &    6 &    2 \\
    pittsburg-bridges-T-OR-D &           77 &    7 &    2 &                             car &         1296 &    6 &    4 \\
               low-res-spect &          398 &  100 &    9 &                     horse-colic &          300 &   25 &    2 \\
                      musk-1 &          357 &  166 &    2 &               heart-switzerland &           92 &   12 &    5 \\
     pittsburg-bridges-REL-L &           77 &    7 &    3 &   oocytes\_trisopterus\_states\_5b &          684 &   32 &    3 \\
               breast-cancer &          215 &    9 &    2 &            congressional-voting &          326 &   16 &    2 \\
                    spambase &         3451 &   57 &    2 &                 acute-nephritis &           90 &    6 &    2 \\
                        iris &          113 &    4 &    3 &                 credit-approval &          518 &   15 &    2 \\
                     thyroid &         3772 &   21 &    3 &                     hill-valley &          606 &  100 &    2 \\
                mammographic &          721 &    5 &    2 &   oocytes\_merluccius\_nucleus\_4d &          767 &   41 &    2 \\
           ilpd-indian-liver &          437 &    9 &    2 &                           seeds &          158 &    7 &    3 \\
                       blood &          561 &    4 &    2 &                           ozone &         1902 &   72 &    2 \\
              waveform-noise &         3750 &   40 &    3 &                           magic &        14265 &   10 &    2 \\
               statlog-heart &          203 &   13 &    2 &                   statlog-image &         1733 &   18 &    7 \\
      pittsburg-bridges-TYPE &           79 &    7 &    6 &                  cylinder-bands &          384 &   35 &    2 \\
              echocardiogram &           98 &   10 &    2 &                     lung-cancer &           24 &   56 &    3 \\
                       flags &          146 &   28 &    8 &                     dermatology &          275 &   34 &    6 \\
                      letter &        15000 &   16 &   26 &       cardiotocography-10clases &         1595 &   21 &   10 \\
                         zoo &           76 &   16 &    7 &                 heart-cleveland &          227 &   13 &    5 \\
                       ecoli &          252 &    7 &    8 &               haberman-survival &          230 &    3 &    2 \\
                       yeast &         1113 &    8 &   10 &                   balance-scale &          469 &    4 &    3 \\
                  hayes-roth &          132 &    3 &    3 &                            wine &          134 &   13 &    3 \\
                      libras &          270 &   90 &   15 &                       miniboone &        97548 &   50 &    2 \\
     breast-cancer-wisc-prog &          149 &   33 &    2 &                       hepatitis &          116 &   19 &    2 \\
                       glass &          161 &    9 &    6 &                  post-operative &           68 &    8 &    3 \\
\bottomrule
\end{tabular}

\end{adjustbox}}
\end{table*}

\end{document}